\pdfoutput=1
\documentclass[11pt]{article}

\usepackage[final]{acl}

\usepackage{times}
\usepackage{latexsym}

\usepackage[T1]{fontenc}
\usepackage[utf8]{inputenc}
\usepackage{microtype}
\usepackage{inconsolata}
\usepackage{graphicx}

\usepackage{xcolor}
\usepackage{multirow}
\usepackage{longtable}
\usepackage{comment}

\usepackage{tikz}
\usetikzlibrary{calc, positioning, shapes.callouts, shapes.misc}
\usepackage{adjustbox}   
\usepackage{varwidth}    

\usepackage{paralist}
\usepackage{enumitem}

\title{From Emotion to Expression: Theoretical Foundations and Resources for Fear Speech}

\author{
Vigneshwaran Shankaran\textsuperscript{1},
Gabriella Lapesa\textsuperscript{1, 2} \and 
Claudia Wagner\textsuperscript{1, 3, 4}\\
\textsuperscript{1}GESIS - Leibniz Institute for the Social Sciences\\ 
\textsuperscript{2}Heinrich-Heine University\\  
\textsuperscript{3}RWTH Aachen University\\
\textsuperscript{4}Complexity Science Hub\\
\texttt{first.last@gesis.org}
}

\begin{document}

\maketitle

\begin{abstract} \label{sec:abstract}

Few forces rival fear in their ability to mobilize societies, distort communication, and reshape collective behavior. In computational linguistics, fear is primarily studied as an emotion, but not as a distinct form of speech. Fear speech content is widespread and growing, and often outperforms hate-speech content in reach and engagement because it appears "civiler" and evades moderation. Yet the computational study of fear speech remains fragmented and under-resourced. This can be understood by recognizing that fear speech is a phenomenon shaped by contributions from multiple disciplines. In this paper, we bridge cross-disciplinary perspectives by comparing theories of fear from Psychology, Political science, Communication science, and Linguistics. Building on this, we review existing definitions. We follow up with a survey of datasets from related research areas and propose a taxonomy that consolidates different dimensions of fear for studying fear speech. By reviewing current datasets and defining core concepts, our work offers both theoretical and practical guidance for creating datasets and advancing fear speech research. 

\end{abstract}

\section{Introduction} \label{sec:introduction}

Fear is a fundamental human emotion that shapes decision-making, risk perception, and social interactions. In fact, it is one of the most primal emotions, driving the fundamental \textit{"fight or flight"} response in most living beings \cite{cannon1939wisdom}. Fear plays a crucial role in driving precautionary actions, fostering group cohesion, and reinforcing ideological boundaries \cite{marshall2007psychology}. Unlike emotions such as happiness or surprise, fear has a different kind of profound influence on human behavior, particularly in digital spaces, where it affects engagement dynamics, online discourse, and information diffusion \cite{brady2017emotion}. Despite its significance, fear as a concept has various interpretations across disciplines that study fear either for its property of being an emotion or its applications as a strategic tool. Regardless, this raises the question of its similarities and divergences and how it could be reconciled to be effectively able to capture the different aspects of it.

\definecolor{lightgreen}{RGB}{208,235,180}
\definecolor{lightred}{RGB}{255,204,204}
\definecolor{lightblue}{RGB}{186,218,255}
\definecolor{lightpurple}{RGB}{221,204,255}
\definecolor{lightyellow}{RGB}{255,244,179}
\definecolor{darkgray}{RGB}{80,80,80}
\definecolor{fearorange}{RGB}{255,165,0}

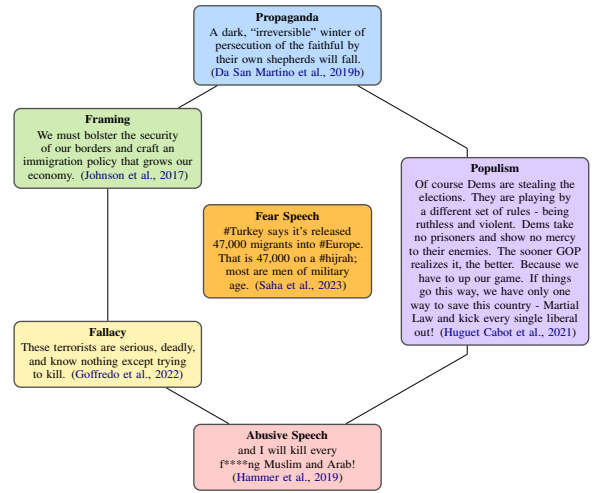
\begin{figure}[t]
\centering
\resizebox{\columnwidth}{!}{

\begin{tikzpicture}[
    box/.style={draw=darkgray, thick, fill=#1, rounded corners=4pt,
                align=center, inner sep=4pt, text width=3.6cm},
    every node/.style={font=\scriptsize}, 
    line/.style={-latex, thick, draw=darkgray!70}
]

\node[box=fearorange!70, text width=3.2cm] (fear)
{\textbf{Fear Speech}\\[1pt]
\#Turkey says it's released 47{,}000 migrants into \#Europe. That is 47{,}000 on a \#hijrah; most are men of military age. \cite{saha2023rise}};

\def\r{4.3cm}
\def\offset{1.9cm} 

\node[box=lightgreen]  (framing)    at (150:\r)
{\textbf{Framing}\\[1pt]
We must bolster the security of our borders and craft an immigration policy that grows our economy. \cite{johnson-etal-2017-leveraging}};

\node[box=lightblue]   (propaganda) at (90:\r)
{\textbf{Propaganda}\\[1pt]
A dark, “irreversible” winter of persecution of the faithful by their own shepherds will fall. \cite{da-san-martino-etal-2019-fine}};

\node[box=lightred]    (abuse)      at (-90:\r)
{\textbf{Abusive Speech}\\[1pt]
and I will kill every f****ng Muslim and Arab! \cite{hammer2019threat}};

\node[box=lightyellow] (fallacy)    at (-150:\r)
{\textbf{Fallacy}\\[1pt]
These terrorists are serious, deadly, and know nothing except trying to kill. \cite{goffredo2022fallacious}};

\node[coordinate] (populism_top) at (30:\r) {};
\node[coordinate] (populism_bottom) at (-30:\r) {};

\draw[-] (framing) -- (propaganda);
\draw[-] (propaganda) -- (populism_top);
\draw[-] (populism_top) -- (populism_bottom);
\draw[-] (populism_bottom) -- (abuse);
\draw[-] (abuse) -- (fallacy);
\draw[-] (fallacy) -- (framing);

\node[box=lightpurple] (populism)   at (0:\r)
{\textbf{Populism}\\[1pt]
Of course Dems are stealing the elections. They are playing by a different set of rules - being ruthless and violent. Dems take no prisoners and show no mercy to their enemies. The sooner GOP realizes it, the better. Because we have to up our game. If things go this way, we have only one way to save this country - Martial Law and kick every single liberal out!
\cite{huguet-cabot-etal-2021-us}};

\end{tikzpicture}

}
\caption{Fear-laden texts are prevalent and have been examined from multiple perspectives.
All examples are sourced from datasets surveyed in this paper.}
\label{fig:examples}
\end{figure}

Fear lies at the core  of the concept of \textit{fear speech}, which has only recently gained traction in computational social science research. In light of the current political climate and the rising prominence of extremist discourse, studying fear speech is crucial, as it can persuade individuals to support or participate in extraordinary measures \cite{wagner2019anxiety}. 
Recent research found that on social media fear speech content is widespread and growing, often outperforming hate-speech content in reach/engagement \cite{saha2023rise}. 

Within the domain of NLP, work on computational argumentation has highlighted the persuasive power of emotionality \cite{benlamine-17} and its toxic potential \cite{ziegenbein-etal-2023-modeling}, focused on emotional language in general and not on specific emotions. \citet{greschner-klinger-2025-fearful} and \citet{quensel-etal-2025-investigating} make a further step annotating discrete emotions (including fear) in arguments and analyzing their impact on persuasion, respectively. Yet, fear has so far been studied as an emotion among others, and it has not yet been put under the interdisciplinary lens we propose in this paper. 

We claim that to study fear speech effectively, theories and resources from multiple disciplines should be considered since fear speech is a cross-disciplinary concept. As often is the case in a data-hungry discipline like NLP, the bottleneck is the lack of datasets that encode specifically the \textit{fear speech signal} (while plenty of annotation is available for fear as an emotion). This is why auditing existing datasets can help identify opportunities for part-reuse, i.e., when fear speech compatible notions have been annotated as a subset of a broader phenomenon, such as \textit{"appeal to fear"} in propaganda research \cite{da-san-martino-etal-2019-fine} or \textit{"immigrant as a threat"} in framing research \cite{mendelsohn-etal-2021-modeling}. This is precisely the goal of the dataset review presented in this paper, whose scope is illustrated in Figure \ref{fig:examples}: starting from the core notion of fear speech devised in the social sciences and encoded in small, high quality datasets (center of the figure), we identify a set of related phenomena (i.e., framing, propaganda, populism, etc.) whose annotations can potentially contribute to the investigation of fear speech, and we conduct a systematic review of datasets and related annotation guidelines. This approach not only leads to an interdisciplinary definition of fear speech, it is also  sustainable because building new annotated resources is costly, so it is important to carefully evaluate what already exists.

\noindent \textbf{Contributions } In this paper, our contributions are threefold. \textbf{First}, we provide a comparative overview of how fear is conceptualized across disciplines such as Political science, Communication science, Psychology, and Linguistics. \textbf{Second}, we review existing definitions of fear speech and argue that one is particularly suitable and expand it in light of the comparative overview of fear. \textbf{Third}, we survey existing datasets that can be used to study fear speech and propose a taxonomy that consolidates and aggregates the different dimensions of fear. Our results show that fear-related constructs are unevenly represented across datasets, with strong concentration in computer science oriented disciplines and significant under representation of social science perspectives.

\section{Theoretical Foundations of Fear} \label{sec:theories}

In this section we discuss fear through various disciplinary lenses. Since our primary interest concerns datasets \textit{(which can potentially be employed for further modeling experiments)} - Political science, Communication science, Psychology and Linguistics offer the most practical foundation for data acquisition and method development.

\subsection{Political Science - Securitization and Elite Rhetoric}

In International Relations and Political Science, securitization theory from the Copenhagen School examines how political elites describe certain issues as existential threats that require emergency actions \cite{buzan1998security}. In this theory, a "securitizing actor," such as a leader, uses a speech act to present a referent object, like the nation or identity, as being in serious danger. When the public accepts this framing, extraordinary measures become justified. Fear, in this context, is not mainly a spontaneous emotional reaction from the public. Instead, it is seen as a discursive tool. Through rhetoric, political actors turn regular political issues into matters of security \cite{balzacq_securitization_2010}. This approach highlights the role of power and intentional manipulation. For example, after 9/11, U.S. leaders used fear of Islamist terrorism to expand executive powers. Securitization shows how defining something as a threat can create collective fear and bypass normal political discussion. Political science also studies how fear shapes electoral politics. Candidates and media often raise public concerns about crime, the economy, or outsiders to influence voters. Campaign ads with ominous visuals are a common example \cite{wagner2019anxiety}. In short, \textbf{Political science views fear as a tool used by elites}. It is created and spread through rhetoric and policy for strategic purposes.

\subsection{Communication Science - Fear Appeals and Persuasion}

In communication research, especially in health and risk communication, fear is often studied as a tool of persuasion. Early models like Protection Motivation Theory (PMT) \cite{prentice1986protection} and the Extended Parallel Process Model (EPPM) \cite{witte1992putting} explain how people respond to fear-based messages. PMT suggests that people make two types of judgments: how serious and likely a threat is, and whether they feel capable of handling it. EPPM builds on this idea. If a message makes people feel both threatened and able to act, they are more likely to take protective steps. If they feel threatened but helpless, they may deny the message or avoid it instead \cite{witte1992putting}. Meta-analyses show that strong fear appeals with clear solutions are more likely to change attitudes and behavior than weak or unclear messages. However, if the message causes too much fear without offering ways to cope, it can have the opposite effect \cite{witte2000meta}. These theories focus on individual reactions. They explain how people notice, interpret, and respond to fear in messages. Fear appeals are common in public health and advertising. For example, anti-smoking campaigns or vaccine messages often use fear to encourage healthy behavior \cite{witte2000meta}. Overall, \textbf{in Communication science, fear is treated as a tool that could be used to drive either beneficial or harmful action}. It is not considered manipulative if it is used with care and includes helpful information.

\subsection{Psychology - Emotion and Appraisal}

Psychologists see fear as a basic emotion tied to survival. It arises when people sense danger and leads to physical changes like increased heart rate or readiness to flee \cite{cannon1939wisdom}. According to appraisal theories, fear happens when someone judges a situation as harmful to their well-being or self-image. Fear can be a short-term state or a longer-term trait like anxiety \cite{smith1990emotion}. Research shows that fear and anxiety increase alertness. People become more cautious, seek out information, and pay closer attention to messages \cite{wagner2019anxiety}. Depending on the situation, fear can lead to helpful actions like problem-solving, or harmful responses like panic and avoidance. Communication science's PMT (theory) also draws on these ideas by including how people judge threats and their ability to handle them. Psychologists also study how people learn fear, such as through classical conditioning in phobias or PTSD (Post-Traumatic Stress Disorder) \cite{rachman1991neo}. Terror management theory adds that fear of death can shape a person’s beliefs and worldview \cite{greenberg1986causes}. Overall, \textbf{Psychology focuses on how people experience and manage fear from the inside}. It often uses tools like anxiety scales or physiological tests to measure fear. 

\subsection{Linguistics - Metaphor, Discourse and Framing}

Linguists and communication scholars study how fear is expressed through language and how the way we talk about issues can shape public opinion. In cognitive linguistics, abstract concepts like fear are often explained using metaphors or physical descriptions \cite{lakoff1980metaphorical}. For example, in French, fear is sometimes described with bodily images like \textit{"turning pale with fear"(pâlir de peur)} or as a growing force, such as \textit{"the fear was growing"(la peur grandissait)} \cite{strobel2015linguistic}. These examples show how people think and talk about fear either as something internal or as something that acts on them. Discourse analysts look at how public language, such as in media or politics, presents certain topics in fearful terms. Phrases like \textit{"flood of immigrants"} or \textit{"war on drugs"} highlight threats while ignoring other views. Researchers analyze who uses such language and how listeners might understand it. This field shows that fear is not just a feeling. It is also built and shared through words, stories, and cultural references. \textbf{Linguists focus on how fear is communicated, what people say, how they say it, and what those choices reveal about social attitudes.} 

\subsection{Comparative Overview: Theories of Fear}

\begin{table*}[h]
	\centering

    \resizebox{\textwidth}{!}{
	\begin{tabular}{p{0.2\linewidth}|p{0.2\linewidth}|p{0.3\linewidth}|p{0.3\linewidth}}
		\hline
		\textbf{Theoretical Approach}                & \textbf{Core Constructs}                                                                      & \textbf{Mechanisms}                                                                                                                                                            & \textbf{Contexts/Examples}                                                                                                                              \\ \hline
		\textbf{Securitization (Political Science)}  & \textit{Threat, Security, Referent Object}                                    & Framing an issue as existential threat via speech acts; elicits public concern and legitimizes extraordinary measures.                                                         & National security, emergency politics, immigration debates (e.g. “war on terror”).                                                                  \\ \hline
		\textbf{Fear Appeals / EPPM (Communication)} & \textit{Perceived Severity, Susceptibility, Efficacy}                                         & Fear-inducing messages trigger threat appraisal; if high efficacy, leads to danger control (adaptive action); if low efficacy, leads to fear control (denial).                 & Health campaigns (anti-smoking, vaccines), public safety ads; advertising (product safety claims).                                                      \\ \hline
		\textbf{Protection Motivation (Psychology)}  & \textit{Threat Appraisal (vulnerability/severity), Coping Appraisal (self/response efficacy}) & Individual cognition: perceiving a threat leads to motivated protective behavior if coping appraisal is favorable; otherwise anxiety or avoidance.                             & Health behavior change (exercise, diet), phobia treatment, climate change risk communication.                                                           \\ \hline
		\textbf{Emotional Appraisal (Psychology)}    & \textit{Emotion (fear/anxiety), Arousal, Cognition}                                           & Fear (a high-arousal negative emotion) emerges from appraisal of low control/danger; it increases vigilance and information seeking but can cause withdrawal or risk aversion. & Personal reactions to news (e.g. fear of crime), therapy (anxiety disorders), media effects on audiences.                                               \\ \hline
		\textbf{Framing \& Metaphor (Linguistics)}   & \textit{Language, Metaphor, Semantics, Narrative Frame}                                       & Linguistic expressions (metaphors like “fear spreads” or “wall of refugees”) shape how audiences conceptualize threats; discourse constructs who/what to fear.       & Media/news discourse (framing immigration as a “wave”), political speeches using metaphor (“this virus is an enemy”), public debate narratives. \\ \hline
	\end{tabular}
    }
    \caption{Comparative Overview: Theories of Fear}
    \label{tab:overview}
\end{table*}

The discipline-specific perspectives outlined in the previous sections can be combined into a multilevel overview for analyzing fear in public discourse.

\begin{itemize}[leftmargin=*]
    \item At the \textbf{Source level}, fear often originates with actors (leaders, media) who select language and frames to present events as threatening. Here, political and linguistic theories intersect: an elite’s speech (securitizing act) or a media headline uses specific metaphors/frames to highlight danger.

    \item At the \textbf{Message level}, communication theory applies: the discourse (e.g., speech, article, social media post) is essentially a fear appeal, characterized by severity/susceptibility cues and possibly efficacy information. One can analyze messages for fear-inducing content (e.g., how many threat words, imagery) and assess whether coping information is provided.

    \item At the \textbf{Reception level}, psychological processes kick in: audiences interpret and respond to these fear messages based on personal traits (trait anxiety), perceived control, and group identities. For instance, if a news story frames a disease as uncontrollable, fearful viewers may experience high anxiety and either panic or seek information (as theory suggests). Public opinion surveys or experiments can test how different fear frames affect people’s attitudes (consistent with political psychology research). 

    \item Finally, at the \textbf{Outcome level}, fear’s impact can be behavioral or political: individuals may support policies (e.g. security measures) they perceive as protective (political science outcome), or adopt personal health behaviors (communication outcome), or they may retreat in denial if efficacy is missing.
\end{itemize}

In practical terms, this comparative overview suggests methodologies for empirical analysis: content analysis of speeches and social media can quantify fear-laden language (drawing on linguistics), discourse analysis can reveal securitizing moves (political science) and metaphoric patterns (linguistics), and surveys/experiments can measure individuals’ fear responses and resulting behaviors (psychology/communication). For example, a study of political speeches might code for “security” frames and then survey listeners’ anxiety levels and policy support to test the whole pathway from discourse to cognition to behavior. Similarly, health communication researchers could analyze news articles or social media about a pandemic (fear frames vs efficacy) and track changes in public precautionary actions. Together, this overview treats fear in public discourse as a process: actors use language (conceptualized and measured by linguistics and political science) to generate emotional cues, which audiences process (per psychological and communication models) and then act upon. By mapping core constructs across disciplines (as in Table \ref{tab:overview}) and tracing them from source to outcome, researchers can capture both the social construction of fear and its psychological effects.

\section{Contextualizing Fear Speech} \label{sec:fearspeech}

Research on fear speech is fairly in its nascent stages and is yet to attract mainstream attention. As with any growing line of research there exist conflicting definitions of fear speech. We have identified three works that explicitly study fear speech and provide two similar yet distinct definitions.

\begin{itemize}[leftmargin=*]
    \item \citet{saha2023rise} adopts the definition by \citet{buyse2014words} from Human Rights studies where fear speech is defined as “an expression
    aimed at instilling (existential) fear of a target (ethnic or religious) group”. \footnote{We cite the latest work of Fear speech by Saha but we acknowledge that the definition was first introduced to the computational research community in \cite{saha2021short}.}
    
    \item \citet{greipl_you_2024} defines fear speech as "any deliberate communicative act that explicitly or implicitly portrays a particular entity, e.g., a group or an institution, as inherently and/or imminently harmful on a cultural, societal, or existential level." 
\end{itemize}

We integrate these two definitions with the broader theoretical perspectives on how fear is generated and utilized in discourse. This expanded view, drawing from disciplines like Political Science, Communication Science, Psychology and Linguistics, sees fear speech as a more complex strategic communicative process. Actors, such as political elites or media outlets, often employ specific linguistic tools like securitizing frames (\textit{"war on terror"}) or evocative metaphors (\textit{"flood of immigrants"}) to construct certain issues, groups, or events as existential threats. These messages frequently function as "fear appeals", designed to influence public perception, legitimize extraordinary measures, or motivate particular behaviors. The ultimate impact of such speech is not uniform, often depending on factors like the audience's psychological processing and whether the message includes information on how to cope with the presented threat (efficacy).

Building upon the definition from \citet{greipl_you_2024} and \citet{saha2023rise}, we define \textbf{fear speech} as "a deliberate communicative act that, through the strategic use of language, rhetoric, and framing, explicitly or implicitly portrays a targeted entity (such as a group, institution, or referent object like the nation or an identity) as an inherent or imminent threat on a cultural, societal, or existential level. The primary aim of the communicative act is to evoke fear or widespread concern in an audience, often to influence their attitudes, justify specific actions, or shape public policy."

We break the definition into three components, which we will use in our dataset analysis:

\begin{itemize}[leftmargin=*]
\item \textbf{C1:} Deliberate communicative acts that use strategic language, rhetoric, and framing.
\item \textbf{C2:} Portrayal of a targeted entity as an inherent or imminent threat at cultural, societal, or existential levels.
\item \textbf{C3:} Primary aim of evoking fear or concern to influence attitudes, justify actions, or shape policy.
\end{itemize}

\paragraph{Fear speech and linguistic theory} We return to linguistic theory, this time with a specific focus on fear speech, motivated by our interest in advancing its study within computational linguistics and exploring how existing linguistic frameworks can inform future modeling approaches for fear speech. Fear speech can also be examined through the lens of linguistic theory, particularly Speech Act Theory \cite{austin1975things, searle1969speech}. Speech act theory offers a useful framework for distinguishing between the literal content of an utterance (locutionary force), the communicative intention of it (illocutionary force), and its potential impact on recipients (perlocutionary force). From this perspective, fear speech is best understood in terms of its illocutionary force, that is, the communicative intention to evoke fear rather than merely to describe a threatening situation (the literal meaning). From a measurement perspective, the literal meaning and the communicative intention of a fear speech instance are definitely less challenging than the perlocutionary effects.  Whether fear is actually induced in the audience, are considerably more difficult to observe and can typically only be approximated through annotation or by analyzing reactions in subsequent discourse.

A further relevant linguistic dimension concerns presupposition. The  presupposition of a statement is the knowledge that is assumed to be shared by the speaker and the receiver of the message \cite{levinson1983pragmatics, beaver1997presupposition}. Such assumption of shared beliefs can be a very powerful communication trick. Consider the use of \textit{again} in the utterance \textit{“Refugees are going to destroy our country again”}: it presupposes that refugees have already destroyed our country in the past  without explicitly arguing for it. Indeed, fear speech often relies on presupposition triggers like \textit{again} that present contested assumptions as given, thereby reinforcing threat narratives implicitly. Prior work on political discourse has shown that such presuppositions play an important role in shaping audience interpretations, casting the use of presupposition triggers like \textit{again} as a form of pragmatic framing \cite{yu-2022-dozens}.

\paragraph{Relation to Hate speech} Fear speech and Hate speech may share certain surface characteristics but we conceptualize fear speech as a broader and analytically distinct phenomenon. In particular, fear speech is not inherently tied to hostility toward a \textbf{protected group}. For instance, public health campaigns such as anti-smoking or anti-drunk-driving messages frequently rely on fear appeals to influence behavior, yet they do not constitute hate speech. While some instances of hate speech may draw on fear-inducing narratives, this overlap alone is insufficient to justify treating the two phenomena within the same analytical framework. We refer to hate speech primarily because its research trajectory illustrates how sustained research attention can emerge in response to strong societal relevance. We argue that fear speech warrants similar attention, especially given the increasingly fear-driven nature of contemporary political and media discourse.

\section{Resource Landscape for Fear Speech}

To foster research on the topic of fear speech, we audit existing resources about it. We conduct a survey on the resource landscape of fear speech from a multi-disciplinary perspective. We use a novel AI-supported prompt based paper discovery method rather than traditional keyword-based literature search methods; this  helps us to address the vocabulary mismatch that may hinder cross-disciplinary literature reviews \cite{schneider-matthes-2024-conversational}.

\subsection{Search Methodology}

\noindent \textbf{Paper Discovery} To systematically discover candidate papers, we employ Allen AI’s Ai2 Paper Finder \textit{(Now part of Ai2 Asta)}\footnote{This tool decomposes prompts into relevant components, searches for papers, follows citation trails, evaluates results for relevance, conducts follow-up queries as needed, and then presents a curated list of papers \cite{AI2PaperFinder}.}. Since prompts are the important element of the process, our overall strategy for designing prompts expands the core concept of \textit{“datasets for fear speech”} and breaking it into component parts, related concepts, and relevant academic contexts. We developed prompts starting from the concept of fear speech and broadened the discovery range by creating prompts for framing, propaganda, emotion appeal, Political science and Communication science. Using this strategy, we ended up with a total of 27 prompts, which together resulted in 810 candidate papers for further screening (Refer to Appendix \ref{app:paper_prompts}). 
To check if the tool was functioning well, we looked at the candidate paper pool to confirm that the results included 3 "core" fear speech papers and 9 prominent papers from propaganda and framing studies that we already identified as relevant using keyword search on Google Scholar.

\paragraph{Categorization Taxonomy} Since we collect datasets from related phenomena, we come across datasets with different labels with similar meanings and definitions. To standardize them, we develop a taxonomy to consolidate and aggregate the different dimensions of fear (Refer to Appendix \ref{app:taxonomy}). The taxonomy is developed to encompass resources that cover at least one dimension of fear from one of these phenomena: Fear Speech, Propaganda, Framing, Populism, Fallacy and Abusive Speech. However, given the heterogeneity of sources and contexts, we cannot assume that each label fully reflects all components described in the taxonomy. We therefore indicate, in parentheses, the specific components that might apply on a case-by-case basis. We omit datasets that are created for the purpose of emotion detection since that is not of interest to studying fear speech.

\paragraph{Paper Screening} Once the candidate papers are identified, the next step is to identify the final set of papers with datasets that could be directly or indirectly used for fear speech research. To support this, we define two buckets of papers to structure our selection process:

\begin{itemize}[leftmargin=*]
    \item \textbf{B1-Fear Speech:} Resources that directly focus on identifying, labeling, or analyzing fear speech.
    
    \item \textbf{B2-Adjacent Phenomena:} Resources that do not explicitly address fear speech but are instrumental for studying fear through related phenomena like framing, propaganda, populism, fallacies and abusive speech which are the top contenders for encoding fear.
\end{itemize}

To systematically review papers and assign them to a suitable bucket, we develop guidelines that align with the fear speech definition presented in Section \ref{sec:fearspeech} (Refer to Appendix \ref{app:paper_screening}). To make the bucketing process robust, we use the components of fear speech. Bucket \textbf{B1} ideally aims to capture all three components directly, whereas bucket \textbf{B2} is intended to capture at least one of the components by focusing on adjacent phenomena that share similar emotional strategies. To streamline the paper selection process, we utilize ASReview, an active learning tool that supports researchers in narrowing down relevant papers.\footnote{ASReview leverages the titles and abstracts of candidate papers, incorporating real-time human input to train machine learning models continuously in the background. This iterative process, referred to by the tool’s creators as a "researcher-in-the-loop" approach, enhances both efficiency and accuracy in identifying relevant literature \cite{van2021open}.} The paper selection was carried out by the first author. To ensure the robustness and quality of the process, we implemented a validation step. After identifying 30 resource-related papers, a stratified sample of 30 papers (including these and others reviewed up to that point) were independently evaluated by the second author and achieved 100 percent consensus. The stopping criterion for ASReview was set at \textbf{50 consecutive non-relevant papers}. In total, \textbf{317 papers} were reviewed, resulting in \textbf{40 papers} that support our research objectives on fear speech. However, during the time of information extraction, two papers were removed because we identified that they did not conform to our research objectives. A complete list of all the \textbf{37 resources} relevant to studying fear speech is provided in Appendix \ref{app:datasets}.


\subsection{Current Landscape}

To map the current landscape of resources on fear-related discourse, we conducted a structured analysis of the existing datasets collected for this study. Our aim is to characterize how fear and related constructs are currently represented and operationalized across research communities. We frame our investigation around four research questions (RQs) that  address disciplinary orientations, data practices, conceptualizations, and dataset reusability.

\paragraph{RQ1: What are the disciplinary orientations of existing studies on fear-related discourse?}
We first examine the disciplinary orientation of each paper. Each dataset was assigned to a research community based on the venue of publication, while for preprints we used the primary subject area indicated by the authors. Our corpus spans \textit{11 distinct research communities}. Of the \textit{37 datasets} we found, \textit{27} of them are contributed by \textit{Computational Linguistics \textbf{(CL)}} venues. The remaining communities - \textit{Computer Science \textbf{(CS)}}, \textit{Machine Learning \textbf{(ML)}}, \textit{Artificial Intelligence \textbf{(AI)}}, \textit{Information Retrieval \textbf{(IR)}}, \textit{Web Mining \textbf{(WM)}}, \textit{Social \& Information Networks \textbf{(SI)}}, \textit{Data Processing \textbf{(DP)}}, \textit{Data Accessibility \textbf{(DA)}}, \textit{Social Science \textbf{(SocSci)}} and \textit{Political Science \textbf{(PolSci)}} each contribute one dataset. When aggregated, \textit{32} out of 37 datasets can be traced to \textbf{computer science–related venues}, reflecting a strong disciplinary concentration and limited engagement from the social sciences.

\begin{figure}[t]
  \includegraphics[width=\columnwidth]{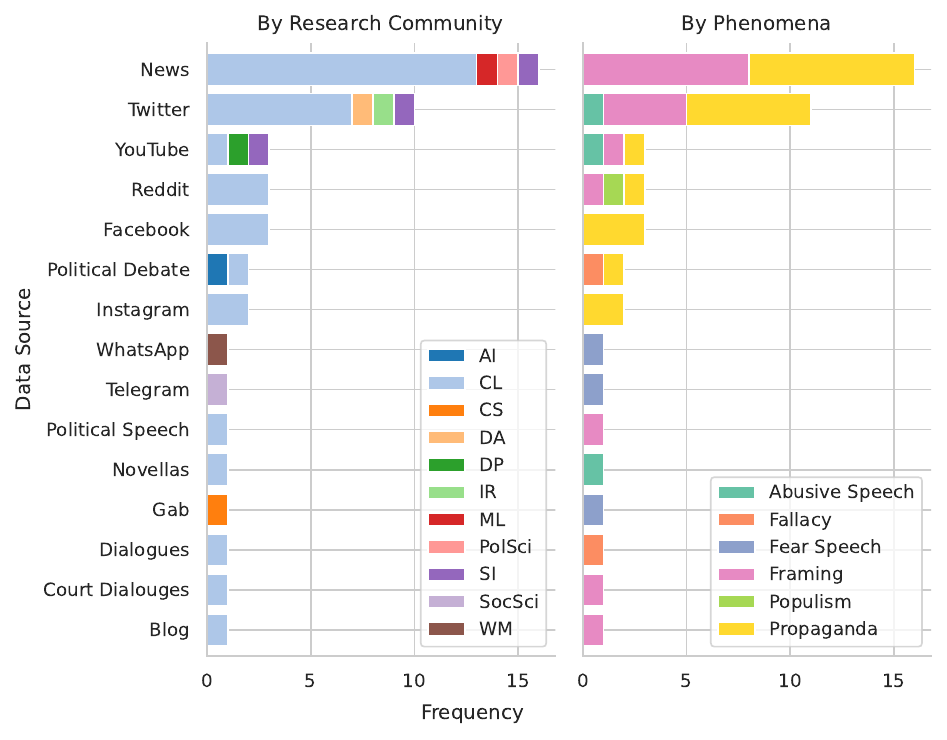}
  \caption{Distribution of research communities and phenomena by data source.}
  \label{fig:datasource}
\end{figure}

\begin{figure}[t]
  \includegraphics[width=\columnwidth]{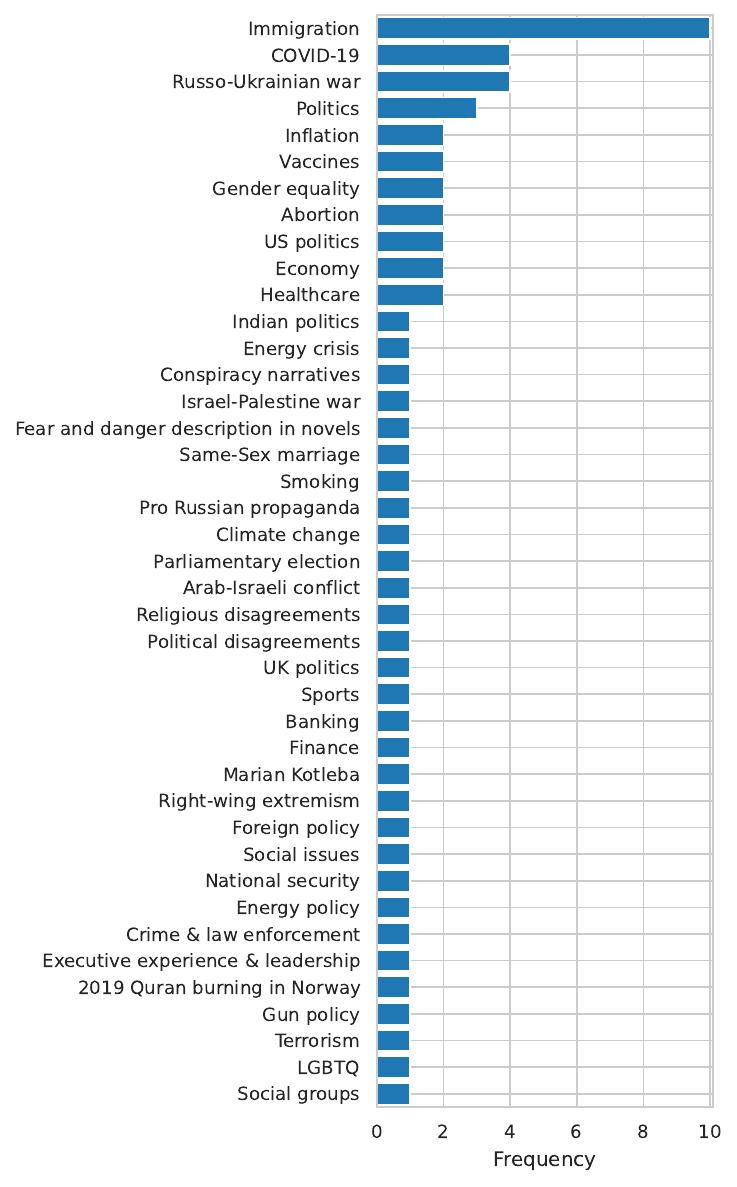}
 \caption{Distribution of topics across datasets}
  \label{fig:topics}
\end{figure}

\paragraph{RQ2: What are the primary data sources used to study fear-related discourse?}
We next coded the data sources underlying each dataset. As shown in Figure \ref{fig:datasource}, \textit{news media} and \textit{Twitter} dominate the landscape, together accounting for \textit{26 datasets}. Interestingly, datasets explicitly designed to capture fear speech were instead sourced from personal or less-moderated communication channels such as \textit{WhatsApp, Telegram, and Gab} -- platforms often associated with private or ideologically charged discourse.
All datasets contained textual data; two included images, two were audiovisual, and one was multi-modal. In terms of language coverage, \textit{English} overwhelmingly dominates (\textit{23} datasets), followed by \textit{German (6)}, \textit{Arabic (5)}, and \textit{Urdu (2)}, with the remainder spanning \textit{Hindi, Czech, French, Italian, Polish, Russian, North Macedonian, Portuguese, Slovak, and Norwegian} each represented once. Figure \ref{fig:topics} illustrates the topical distribution across datasets. Among those with identifiable topics,\footnote{12 datasets  had no explicit topic information, either because it was not provided or because the data was source oriented (e.g., Reddit communities; curated news media lists).} immigration appears most frequently \textit{(10 datasets)}, followed by politics including US, Indian, and UK politics in \textit{8 datasets}. Other topics include \textit{climate change}, \textit{wars (Israel-Palestine and Russo-Ukrainian)}, \textit{sports}, and \textit{finance}, suggesting a moderate topical diversity despite strong thematic clustering around political and social conflict.

\begin{figure}[h]
  \includegraphics[width=\columnwidth]{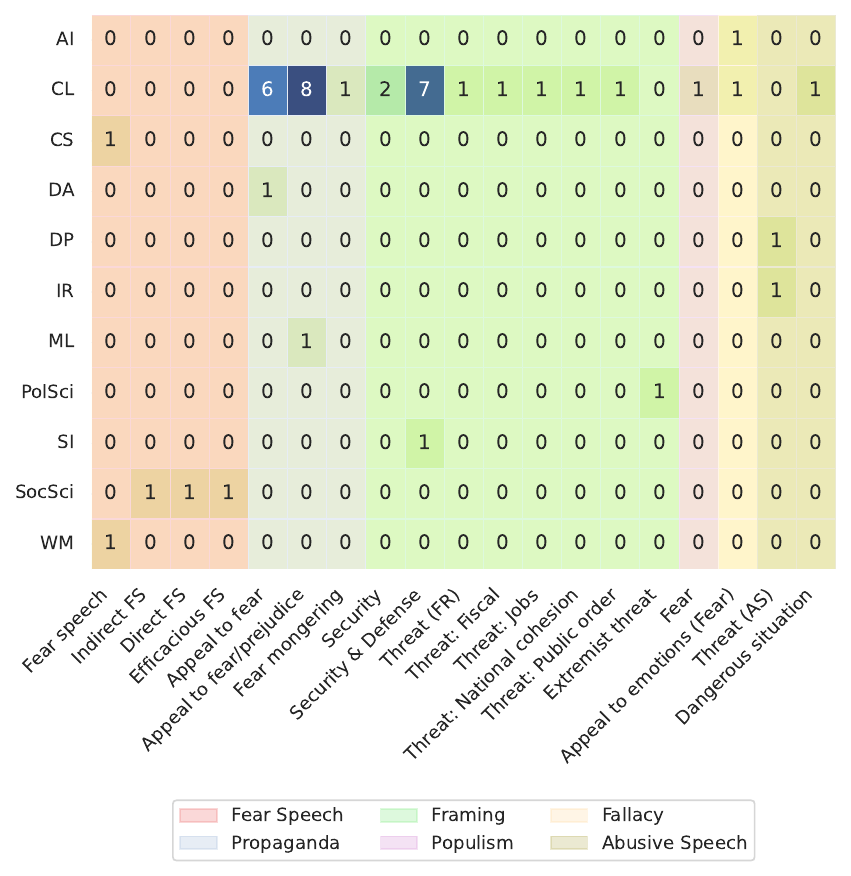}
 \caption{Distribution of fear related labels across research communities grouped by phenomena.}
  \label{fig:fear}
\end{figure}

\paragraph{RQ3: How are fear-related phenomena conceptualized and labeled across datasets?}
We identified 19 distinct labels across all datasets and aligned them with our proposed taxonomy of fear dimensions (Figure~\ref{fig:fear}). The most frequent conceptualizations correspond to propaganda and framing, underscoring the dominance of these perspectives in studying fear discourse. Notably, labels \textit{"Appeal to fear"} and \textit{"Appeal to fear/prejudice"} even though a common propaganda technique has minor difference with the latter using emotional appeal to capitalize on people's existing bias and stereotypes. Several fallacy focused datasets included the generic label \textit{“Appeal to emotion”}, but were excluded in absence of fear-specific sub labels. As a result, \textit{fallacy} and \textit{populism} emerge as the most underrepresented phenomena in the current dataset landscape.


\paragraph{RQ4: To what extent are existing datasets accessible and reusable?}
Finally, to assess the potential for resource part reuse, we evaluated each dataset’s accessibility, annotation transparency, and licensing status. Of the 37 datasets, 26 provided some form of access mechanism most commonly via \textit{direct link (23)}, followed by \textit{registration portal (2)} or \textit{contact form (1)}. However, only 21 of these links remained active at the time of our auditing, raising concerns about long term resource persistence. Moreover, only 13 datasets included annotation guidelines, reflecting limited documentation of labeling practices. Among the 33 manually annotated datasets, only 22 reported inter-annotator agreement scores and just 7 mention information of the annotators. Reusability is further constrained by incomplete reporting of corpus size. For 7 datasets, the number of fear related instances was not specified in the paper; for 4 others, it was determined manually by direct inspection of the data. These findings highlight significant variability in transparency and documentation practices, which  impede reproducibility and reuse across the fear discourse research landscape.

\section{Discussion}
\label{sec:discussion}

The motivation behind this study lies in a simple observation that despite fear's profound role in shaping public discourse, computational linguistics lacks a coherent direction or resource base for studying it. While hate speech and toxicity have inspired large-scale datasets and shared tasks, fear speech remains theoretically diffuse and empirically under-sourced. Our cross-disciplinary approach and resource mapping aim to establish fear as a valid construct for computational modeling grounded in theory yet operationally feasible.

From a computational perspective, this study provides a strategic starting point for modularizing fear. The comparative overview in Section \ref{sec:theories} shows that fear manifests through identifiable linguistic and rhetorical cues such as metaphors, threat framing and emotional appeals which can be detected with existing NLP techniques. Future work can decompose fear into modular signals aligned with our definition components: strategic language, portrayal of threat and intent to evoke fear. This modular view enables researchers to reuse and combine methods like propaganda detection and framing while keeping theoretical coherence.

For the broader computational linguistics community, the value of our resource analysis lies in guidance. Understanding where fear-related signals already exist enables more targeted corpus design, whether through extending current resources, refining annotation schemes, or addressing gaps in underrepresented contexts such as everyday online discourse, local news, and crisis communication. Expanding linguistic coverage beyond English and German and improving annotation transparency through open guidelines and agreement reporting directly address the accessibility gaps identified in our audit. Our actively maintained repository\footnote{\url{https://fearspeechdata.com/} \textit{(A successful pull request to this repository would expand the catalog of datasets)}} continuously updated with new relevant datasets by the community, will serve as shared infrastructure for the emerging fear speech research.

Looking ahead, connecting theoretical insight with scalable NLP practice can advance fear speech as a distinct computational task. Future fear speech research can integrate modular signal detection with dynamic, community driven datasets that evolve alongside social discourse. Such an agenda not only advances fear speech as a computational task but also strengthens our capacity to study how emotions structure public communication at scale. Ultimately, our findings invite the computational linguistics community to view fear not only as a marginal emotional category among established phenomena, but also as a core analytic dimension.

\section{Conclusion} \label{sec:conclusion}

This paper bridges theoretical perspectives and empirical resources to position fear speech as a distinct yet integrative phenomenon. By mapping how fear is conceptualized across disciplines and auditing datasets from adjacent research areas, we identify both the conceptual gaps and practical opportunities: fear is rarely modeled as a communicative construct, annotation practices are inconsistent, and existing resources remain concentrated within computational rather than interdisciplinary domains.

Our comparative overview shows that fear can be operationalized through modular linguistic and rhetorical signals. The resource landscape we present offers a foundation for informed, continuous, collaborative data development, encouraging the computational linguistics community to extend fear-relevant datasets as shared infrastructure. In doing so, we move towards a research timeline where fear is not treated as a marginal emotional category but as a core analytic dimension that connects emotion, social influence and linguistic form.

\section*{Limitations} \label{sec:limitations}

\paragraph{Dual use} While the goal of this work is to build a solid, theory-based  empirical foundation for the understanding (and modeling) of fear speech, we cannot exclude that the our dataset survey may be used to develop systems which use fear to manipulate users. 

\paragraph{Scope of the investigated phenomena} Our overview of interdisciplinary phenomena and constructs is centered on aspects that have to do with the use of fear speech as a \textit{"a deliberate communicative act"}. That is, we focus on the side of the \textit{"producer"} of the fear speech and on the communicative strategies/motives that lead to its use. We acknowledge that a whole set of other constructs from social-psychology can in future work be integrated into the study of fear speech: these are phenomena that define which type of \textit{"audience"} is more susceptible to fear speech. This could be the case for example for individuals: with a preference for values such as "security" (Human Value Theory, \citet{Schwartz2004}); with a strong loading on moral foundations such as "care/harm" (Moral Foundation Theory, \citet{Haidt2004}); cultural dimensions such as "uncertainty avoidance" (Hofstede's theory of cultural dimensions, \citet{hofstede2001culture}); or with a regulatory focus oriented to "prevention" (Regulatory Focus Theory, \citet{doi:https://doi.org/10.1002/9781118900772.etrds0279}).

\paragraph{Theoretical Scope and Exclusion of Methodological Analysis} This paper is about theories and datasets, and is not concerned with modeling choices or specific methods. For example, most political science, psychology and communication science works rely on qualitative screening, dictionary-based approaches and off-the-shelf tools like LIWC, while NLP research adopts models that are more powerful but increasingly less interpretable. While we acknowledge this to be a very important aspect, an overview of methods and approaches falls out of the scope of this paper and we reserve it for future work.

\paragraph{Scope of the investigated papers} While we highlight a prevalence of computer-science contributions as dataset sources, we cannot exclude that this finding is driven by the fact that papers are labeled by venue and authors from social, political or communication science could be involved as authors.

\paragraph{AI-Assisted Literature Survey} The literature review process was conducted across multiple disciplinary domains that lie outside our primary areas of expertise. As a result, conventional keyword-based search strategies were at times insufficient, as they rely heavily on prior familiarity with domain-specific terminology. This limitation became evident, for example, in our delayed identification of the concept of securitization, which only emerged after consultation with a political scientist. In such contexts, semantic and AI-assisted search methods can offer substantial advantages by enabling the discovery of relevant work that may not be captured through standard keyword queries \cite{schneider-matthes-2024-conversational}. While a systematic evaluation of these methods against established literature review standards would constitute a valuable contribution to research on review methodologies, such an analysis lies beyond the scope of the present study. Accordingly, the use of AI-assisted search should be interpreted as a pragmatic support tool rather than a formally validated alternative to conventional systematic review procedures.

\section*{Acknowledgments} \label{sec:acknowledgments}

We would like to thank Luisa Kotthoff and Annika Herbertz for their support in the collection of meta data of the surveyed datasets.

\bibliography{anthology, custom, datasets}

\appendix
\label{sec:appendix}

\section{Prompts}
\label{app:paper_prompts}
Table \ref{tab:prompts} lists the prompts used to discover candidate  papers.

\section{Information Extraction}
\label{app:information_extraction}
We trained and employed two student assistants to extract metadata from the datasets used in the selected papers. Our procedure adapts the systematic literature database developed by \cite{romberg-etal-2025-towards} to guide future research on fear speech. We began by collecting fundamental information about each dataset: the \textbf{dataset name}, and its \textbf{availability} (\textit{online}, \textit{upon request}, \textit{upon registration}, or \textit{-}). For datasets available online, we additionally recorded the \textbf{resource link} and its \textbf{status} (\textit{accessible}, \textit{not accessible} or \textit{-} as of September 2025). We also noted the \textbf{license}, \textbf{annotation type} (\textit{in-house}, \textit{crowd-sourced}, \textit{external}, \textit{automatic}\footnote{Some papers implement novel methods to label data and validate with a gold corpus. An example is in line with the recent advances and trends in CL research, prompting LLMs.}), and, in the case of manual annotation, the \textbf{availability of annotation guidelines} (\textit{Yes} or \textit{-}). For the corresponding papers that introduced the datasets, we recorded the \textbf{title}, \textbf{authors}, \textbf{year of publication}, and \textbf{URL}, along with the targeted \textbf{research community} (based on the publication venue). At the dataset level, we documented the \textbf{size} in terms of \textbf{annotation units}, the \textbf{topic}, the \textbf{modality} (\textit{text}, \textit{image}, \textit{audio-visual} or \textit{multi-modal}), and the \textbf{language}. To align with our focus on fear speech, we included \textbf{label} as presented in the original papers and mapped these to the \textbf{dimension} from our \textit{fear speech taxonomy codes} (Refer to Appendix \ref{app:taxonomy}). For manually annotated datasets, we also tracked inter-annotator agreement through the reported \textbf{IAA score} and \textbf{IAA measure}.

\section{Taxonomy}
\label{app:taxonomy}
Table \ref{tab:taxonomy} describes the taxonomy developed for categorizing fear dimensions.

\section{Paper Screening Guidelines}
\label{app:paper_screening}

\noindent \textbf{Objective: } Your task is to review a list of academic papers and classify each one into one of three categories:
\begin{itemize}
    \item \textbf{B1:} Fear Speech
    \item \textbf{B2:} Adjacent Phenomena
    \item Discard
\end{itemize}

This classification is for a research project on identifying relevant datasets and resource for studying fear speech.

\noindent \textbf{Definition:} We define fear speech as a \textit{deliberate communicative act that, through the strategic use of language, rhetoric, and framing}, explicitly or implicitly \textit{portrays a targeted entity} (such as a group, institution, or referent object like the nation or an identity) \textit{as an inherent or imminent threat on a cultural, societal, or existential level}. Its \textit{primary aim is to evoke fear or widespread concern in an audience, often to influence their attitudes, justify specific actions, or shape public policy.}

Breaking it down into core components:
\begin{itemize}
    \item \textbf{C1:} Deliberate communicative acts that use strategic language, rhetoric, and framing.
    \item \textbf{C2:} Portrayal of a targeted entity as an inherent or imminent threat at cultural, societal, or existential levels.
    \item \textbf{C3:} Primary aim of evoking fear or concern to influence attitudes, justify actions, or shape policy.
\end{itemize}

\noindent \textbf{B1: Fear Speech}
\begin{itemize}
    \item \textbf{Focus:} Papers that directly study the concept of "fear speech" even if their definition of fear speech is not the one defined above.
    \item \textbf{Criteria:} The paper must explicitly mention “fear speech”.
\end{itemize}

\noindent \textbf{B2 - Adjacent Phenomena:}
\begin{itemize}
    \item \textbf{Focus:} Papers that don't use the term "fear speech" but focus on related phenomena like propaganda, framing, populism and fallacy.
    \item \textbf{Criteria:} The paper focuses on at least one of the components: \textbf{Component 1 (strategic language)}, \textbf{Components 2 (portrayal as threat) and 3 (evoking fear)}. It studies the mechanisms that are used in fear speech, even if it labels them differently.
    \item \textbf{Think of it this way:} Could this paper's dataset or method be used to analyze fear speech, even if the authors didn't call it fear speech? If yes, it's likely B2.
\end{itemize}

\noindent \textbf{The Classification Process: A Step-by-Step Workflow}

\noindent \textbf{Step 1: Initial Scan :} Read the Title, Keywords, and Abstract. This is often enough to infer if the paper is introducting a dataset. If you are not able to find the mention of “dataset” in the abstract, CTRL+F “dataset” and skim the highlighted marks area if it communicates the creation of a new dataset.

\noindent \textbf{Step 2: B1 Test :} Ctrl+F and search for \textit{“fear speech”} and \textit{“fearspeech”}. Now examine if the paper introduces a dataset and is explicitly labeled for fear speech. If this is not the case, its not in B1 bucket, proceed further.

\noindent \textbf{Step 3: B2 Test :} Ask these questions to determine if paper belongs to B2. \underline{Note:} Emotion detection resources do not belong in this bucket and must be discarded.
\begin{enumerate}
    \item Does the paper have one of the key words: \textit{“fear”, “anxiety”, “threat”, “security”, “risk”, “crisis”}? \underline{Note:} In case you find similar keywords that are not listed here and could be used to extend it, please note it down and we will resolve it during discussion session.

    \item Does it come under one of the following phenomena: \textit{“framing”, “propaganda”, “populism”, “fallacy”, “crisis communication”, "abusive speech"} ? \underline{Note:} In case you find a similar phenomenon that is not listed here, please note it down and we will resolve it during discussion session.
        
    \item Does the dataset contain one of the following labels or similar labels: \textit{"appeal to fear", "fear mongering", "security". "threat", "crisis", "appeal to emotion"}? \underline{Note:} When there is a mention of Appeal to emotion as in the case of paper from Fallacy studies, choose them only if have "fear" has sub-category else discard.
\end{enumerate}

\noindent \textbf{Step 4: Discard :} If the paper does not fit B1 or B2, classify it as Discard. 
\begin{enumerate}
    \item It could be \textbf{too broad} like \textit{“hate speech”} or \textit{“offensive language”} but \textbf{lacks any dimension of fear, anxiety or threat.} 
    
    \item It could be a \textbf{non-useful threat} (ex: virus, malware in cybersecurity research).
        
    \item \textbf{Method-only studies} that reuse existing datasets without expanding on existing dataset or introducing new datasets or resources.
    
    \item If it’s an \textbf{emotion detection resource.}
\end{enumerate}

\section{List of Datasets}
\label{app:datasets}
Table \ref{tab:datasets} lists the 37 datasets that we identified in our systematic literature search.

\begin{table*}
	[h]
	\centering
	\resizebox{\textwidth}{350pt}{
	\begin{tabular}{l p{0.5\linewidth} p{0.40\linewidth}}
		\hline
		\textbf{Dimension}      & \textbf{Description}                                                                                                                                                                                                                                                                                                                                                                                                                          & \textbf{Components}                                                                                                                                                                 \\
		\hline
		\textbf{Fear Speech}    & \multirow{3}{\linewidth}{Fear speech is a deliberate communicative act that, through the strategic use of language, rhetoric, and framing, explicitly or implicitly portrays a targeted entity as an inherent or imminent threat on a cultural, societal, or existential level. Its primary aim is to evoke fear or widespread concern in an audience, often to influence their attitudes, justify specific actions, or shape public policy.} & C1: Deliberate communicative acts that use strategic language, rhetoric, and framing.                                                                                               \\
		                        &                                                                                                                                                                                                                                                                                                                                                                                                                                               & C2: Portrayal of a targeted entity as an inherent or imminent threat at cultural, societal, or existential levels.                                                                  \\
		                        &                                                                                                                                                                                                                                                                                                                                                                                                                                               & C3: Primary aim of evoking fear or concern to influence attitudes, justify actions, or shape policy.                                                                                \\
		\hline
		\textbf{Propaganda}     &                                                                                                                                                                                                                                                                                                                                                                                                                                               &                                                                                                                                                                                     \\
		Appeal to fear          & \multirow{2}{\linewidth}{Seeking to build support for an idea by instilling anxiety and/or panic in the population. towards an alternative, possibly based on preconceived judgments. \cite{da-san-martino-etal-2020-semeval}}                                                                                                                                                                                                                & C1: A deliberate rhetorical strategy to persuade an audience.                                                                                                                       \\
		                        &                                                                                                                                                                                                                                                                                                                                                                                                                                               & (C2): It typically portrays an group, event, or idea as an imminent or existential danger.                                                                                          \\
		Fear mongering          & \multirow{2}{\linewidth}{Is the text trying to appeal to fear, uncertainty or other threat? \cite{horak2024recognition}}                                                                                                                                                                                                                                                                                                                      & C1: A deliberate rhetorical strategy to persuade an audience.                                                                                                                       \\
		                        &                                                                                                                                                                                                                                                                                                                                                                                                                                               & (C2): It typically portrays an group, event, or idea as an imminent or existential danger.                                                                                          \\
		\hline
		\textbf{Framing}        &                                                                                                                                                                                                                                                                                                                                                                                                                                               &                                                                                                                                                                                     \\
		Security                & \multirow{3}{\linewidth}{Used as a framing label to present issues (e.g., immigration, terrorism) as threats to safety, emphasizing protection or defense}                                                                                                                                                                                                                                                                                    & C1: An intentional framing device.                                                                                                                                                  \\
		                        &                                                                                                                                                                                                                                                                                                                                                                                                                                               & C2: By design it casts issues (for example migration) in terms of threats to safety or the social order.                                                                            \\
		                        &                                                                                                                                                                                                                                                                                                                                                                                                                                               & (C3): Primarily used to prioritize protective measures and shape policy preferences, they commonly aim to induce concern or support for defensive action.                           \\
		Threat                  & \multirow{3}{\linewidth}{Frames an entity, group, or event as posing danger and overlapping with security framing or as a broader construct of fear or crisis.}                                                                                                                                                                                                                                                                               & C1: An intentional framing device.                                                                                                                                                  \\
		                        &                                                                                                                                                                                                                                                                                                                                                                                                                                               & C2: It explicitly depicts an actor or event as dangerous at cultural, societal, or existential levels.                                                                              \\
		                        &                                                                                                                                                                                                                                                                                                                                                                                                                                               & (C3): Primarily used to prioritize protective measures and shape policy preferences, they commonly aim to induce concern or support for defensive action.                           \\
		\hline
		\textbf{Populism}       &                                                                                                                                                                                                                                                                                                                                                                                                                                               &                                                                                                                                                                                     \\
		Crisis rhetoric         & \multirow{3}{\linewidth}{A communication strategy that identifies a "failure" or threatening event and elevates it to the level of an existential crisis. \cite{Moffitt_2015}}                                                                                                                                                                                                                                                                & C1: It is an intentional rhetorical strategy aimed at mobilizing “the people”.                                                                                                      \\
		                        &                                                                                                                                                                                                                                                                                                                                                                                                                                               & C2: It casts the situation as a systemic or existential breakdown, often locating blame in elites or out-groups and portraying those actors as threats to the people’s way of life. \\
		                        &                                                                                                                                                                                                                                                                                                                                                                                                                                               & (C3): It seeks to evoke fear, anger, or urgency among the audience to legitimize the populist leader’s proposed solutions or policies.                                              \\
		\hline
		\textbf{Fallacy}        &                                                                                                                                                                                                                                                                                                                                                                                                                                               &                                                                                                                                                                                     \\
		Appeal to emotion       & \multirow{2}{\linewidth}{Appeal to Emotion Manipulation of the recipient’s emotions in order to win an argument. \cite{jin-etal-2022-logical}}                                                                                                                                                                                                                                                                                                & C1: A deliberate rhetorical strategy to persuade an audience.                                                                                                                       \\
		                        &                                                                                                                                                                                                                                                                                                                                                                                                                                               & (C2): It typically portrays an group, event, or idea as an imminent or existential danger. This applies only if the emotion appealed to is fear.                                    \\
		\hline
		\textbf{Abusive Speech} &                                                                                                                                                                                                                                                                                                                                                                                                                                               &                                                                                                                                                                                     \\
		Threat                  & These have language intended to make the target or a broader group fearful or to feel unsafe. The threats may be personal or general. Maybe explicit or they may generally make the target feel unsafe without a specific threat of direct action. \cite{golbeck2017large}                                                                                                                                                                    & C3: Its immediate objective is to make the target or targets fearful and unsafe.                                                                                                    \\
		\hline
	\end{tabular}
	}
	\caption{Taxonomy for dimensions of fear.}
	\label{tab:taxonomy}
\end{table*}

\begin{table*}[]
\centering
\begin{tabular}{l p{0.85\textwidth}} 
\cline{1-2}
\multicolumn{2}{l}{\textbf{Fear Speech}} \\ 
\cline{1-2}
1     & Papers that present datasets for detecting fear speech. \\
2     & Papers that present datasets for detecting deliberate communicative act that, through the strategic use of language, rhetoric, and framing, explicitly or implicitly portrays a targeted entity (such as a group, institution, or referent object like the nation or an identity) as an inherent or imminent threat on a cultural, societal, or existential level and its primary aim is to evoke fear or widespread concern in an audience, often to influence their attitudes, justify specific actions, or shape public policy. \\
3     & Papers that present datasets for detecting deliberate communicative acts that use strategic language, rhetoric, and framing. \\
4     & Papers that present datasets for detecting portrayal of a targeted entity as an inherent or imminent threat at cultural, societal, or existential levels. \\
5     & Papers that present datasets for detecting text with the primary aim of evoking fear or concern to influence attitudes, justify actions, or shape policy. \\ 
\cline{1-2}
\multicolumn{2}{l}{\textbf{Framing}} \\ 
\cline{1-2}
6-9   & Papers that present datasets for detecting threat-based framing strategies \textit{(in [political speech, media coverage, online discourse])}, where framing is defined as the process of selecting certain aspects of a perceived reality and making them more salient in a communicating text, in order to promote a particular problem definition, causal interpretation, moral evaluation, and/or treatment recommendation. \\
10-13 & Papers that present datasets for detecting strategic framing used to depict existential or cultural threats \textit{(in [political text, media coverage, online discourse])} , where framing is defined as the process of selecting certain aspects of a perceived reality and making them more salient in a communicating text, in order to promote a particular problem definition, causal interpretation, moral evaluation, and/or treatment recommendation. \\
14-17 & Papers that present datasets for detecting framing devices that promote fear \textit{(in [political speech, media coverage, online discourse])}, where framing is defined as the process of selecting certain aspects of a perceived reality and making them more salient in a communicating text, in order to promote a particular problem definition, causal interpretation, moral evaluation, and/or treatment recommendation. \\ 
\cline{1-2}
\multicolumn{2}{l}{\textbf{Propaganda}} \\ 
\cline{1-2}
18    & Papers that present datasets for detecting fear-based propaganda. \\
19    & Papers that present datasets for detecting propaganda techniques, specifically "appeal to fear," "threat invocation," or "whataboutism." \\
20    & Papers that present datasets for identifying propaganda techniques used to portray an out-group as a danger. \\ 
\cline{1-2}
\multicolumn{2}{l}{\textbf{Emotion Appeal}} \\ 
\cline{1-2}
21    & Papers that present datasets for detecting manipulative language strategies that evoke fear or anxiety. \\
22    & Papers that present datasets for detecting logical fallacies driven by emotion like appeal to fear. \\ 
\cline{1-2}
\multicolumn{2}{l}{\textbf{Political Science}} \\ 
\cline{1-2}
23    & Papers on securitization theory that use computational methods to analyze speech acts that frame issues as existential threats. \\
24    & Papers with datasets of parliamentary debates or policy documents annotated for securitizing moves and referent object construction. \\ 
\cline{1-2}
\multicolumn{2}{l}{\textbf{Communication Science}} \\ 
\cline{1-2}
25    & Papers that present datasets for detecting fear-inducing narratives in mass media reporting. \\
26    & Papers that present datasets for detecting fear-based rhetoric during crises. \\
27    & Papers that present datasets for analyzing emergency communication strategies that evoke fear or urgency. \\
\cline{1-2}
\end{tabular}
\caption{Prompts used to identify candidate papers through Allen AI’s AI2 Paper Finder. For the Framing category, each row includes three prompt options, with the values in brackets indicating the chosen option. }
\label{tab:prompts}
\end{table*}

\clearpage
\onecolumn
\begin{longtable}{p{0.15\linewidth} p{0.1\linewidth} p{0.15\linewidth} p{0.1\linewidth} p{0.15\linewidth} p{0.1\linewidth}}

\hline \multicolumn{1}{c}{\textbf{Paper}} & \multicolumn{1}{c}{\textbf{Community}} & \multicolumn{1}{c}{\textbf{Label}} & \multicolumn{1}{c}{\textbf{Dimension}} & \multicolumn{1}{c}{\textbf{Topic}} & \multicolumn{1}{c}{\textbf{Data Source}} \\ \hline 
\endfirsthead

\multicolumn{6}{c}%
{{\bfseries \tablename\ \thetable{} -- continued from previous page}} \\
\hline \multicolumn{1}{c}{\textbf{Paper}} & \multicolumn{1}{c}{\textbf{Community}} & \multicolumn{1}{c}{\textbf{Label}} & \multicolumn{1}{c}{\textbf{Dimension}} & \multicolumn{1}{c}{\textbf{Topic}} & \multicolumn{1}{c}{\textbf{Data Source}} \\ \hline 
\endhead

\hline \multicolumn{6}{r}{{Continued on next page}} \\ \hline
\endfoot

\hline
\caption{Overview of datasets, including the research community
targeted \textit{(CL - Computational Linguistics, CS - Computer Science, ML - Machine Learning,  AI - Artificial Intelligence, IR - Information Retrieval, WM - Web Mining, SI - Social \& Information Networks, DP - Data Processing, DA - Data Accessibility, SocSci - Social Science, PolSci - Political Science)}, the label(s) in the dataset, the assigned dimension of fear according to our taxonomy \textit{(FS:FS - Fear speech, PR:AP - Appeal to fear in Propaganda, PR:FM - Fear mongering in Propaganda, FR:SE - Security framing, FR:TH - Threat framing, PO:CR - Crisis rhetoric in populism, FA:AE - Appeal to emotion in Fallacy, AS:TH - Threat in Abusive speech)}, topic of the dataset and the data source} \label{tab:datasets} \\
\endlastfoot

\citet{saha2021short} & WM & Fear speech & FS:FS & Indian politics & WhatsApp \\ 
\citet{saha2023rise} & CS & Fear speech & FS:FS & - & Gab \\ 
\citet{greipl2024you} & SocSci & Indirect FS, Direct FS, Efficacious FS & FS:FS & COVID-19, Energy Crisis, Inflation, Immigration, Russo-Ukrainian war, Conspiracy Narratives & Telegram  \\ 
\citet{mendelsohn-etal-2021-modeling} & CL & Threat: Jobs, Threat: Public order, Threat: National cohesion, Threat: Fiscal, Security \& Defense & FR:SE, FR:TH & Immigration & Twitter \\ 
\citet{arora2025multi} & CL & Security \& Defense & FR:SE   & Immigration & News \\ 
\citet{huguet-cabot-etal-2021-us} & CL & Fear & PO:CR & Social groups (Immigrants, Refugees, Muslims, Jews, Liberals, Conservatives) & Reddit \\ 
\citet{kaur2025beyond} & CL & Security \& Defense & FR:SE & Israel-Palestine war & News \\ 
\citet{yu-fliethmann-2022-frame} & CL & Security & FR:SE & Immigration & News \\ 
\citet{yu-2023-towards} & CL & Security & FR:SE & Immigration & News \\ 
\citet{zehe2023towards} & CL & Dangerous situation & AS:TH & Fear and danger description in novels & Novellas \\ 
\citet{card-etal-2015-media} & CL & Security \& Defense & FR:SE & Immigration, Same-Sex Marriage, Smoking & News \\ 
\citet{salman-etal-2023-detecting} & CL & Appeal to fear/prejudice & PR:AF & - & Twitter, Facebook, Instagram, YouTube \\ 
\citet{dimitrov-etal-2021-detecting} & CL & Appeal to fear/prejudice & PR:AF & Vaccines, Politics, COVID-19, Gender Equality & Facebook  \\ 
\citet{hasanain-etal-2024-gpt} & CL & Appeal to fear/prejudice & PR:AF  & - & News \\ 
\citet{da-san-martino-etal-2019-fine} & CL & Appeal to fear/prejudice & PR:AF & - & News \\ 
\citet{vijayaraghavan-vosoughi-2022-tweetspin} & CL & Appeal to fear/prejudice & PR:AF & - & Twitter  \\ 
\citet{yu2019experiments} & ML & Appeal to fear/prejudice & PR:AF & - & News \\ 
\citet{alam-etal-2022-overview} & CL & Appeal to fear/prejudice & PR:AF & - & Twitter \\ 
\citet{baisa-etal-2019-benchmark} & CL & Fear mongering & PR:FM & Pro Russian Propaganda & News \\ 
\citet{piskorski-etal-2023-multilingual} & CL & Security \& Defense, Appeal to fear/prejudice & PR:AF & COVID-19, Abortion, Immigration, Russo-Ukrainian war, Climate Change, Parliamentary Election & News \\ 
\citet{hammer2019threat} & DP & Threat & AS:TH & Arab-Israeli conflict, Religious Disagreements, Political Disagreements & YouTube  \\ 
\citet{hasanain-etal-2023-araieval} & CL & Appeal to fear & PR:AF & - & Twitter, News \\ 
\citet{dimitrov-etal-2024-semeval} & CL & Appeal to fear/prejudice & PR:AF & Politics, Vaccines, COVID-19, Geneder Equality, Russo-Ukrainian war & Facebook, Instagram \\ 
\citet{amjad2022overview} & IR & Threat & AS:TH & - & Twitter \\ 
\citet{da-san-martino-etal-2019-findings} & CL & Appeal to fear & PR:AF & - & News \\ 
\citet{da-san-martino-etal-2020-semeval} & CL & Appeal to fear & PR:AF & - & News \\ 
\citet{iyer2019unsupervised} & CL & Threat & FR:TH & US politics & Reddit, Court Dialouges, Political Speech, Blog \\ 
\citet{balalau-horincar-2021-stage} & CL & Appeal to fear & PR:AF & UK politics, US politics & Reddit \\ 
\citet{maarouf-etal-2024-hqp} & CL & Appeal to fear & PR:AF & Russo-Ukrainian war & Twitter \\ 
\citet{almotairy2024dataset} & DA & Appeal to fear & PR:AF & Sports, Banking & Twitter \\ 
\citet{daffara2025generalizability} & CL & Security \& Defense & FR:SE & Politics, Economy, Finance & News \\ 
\citet{mieleszczenko2025unraveling} & CL & Appeal to emotions (fear) & FA:AE & - & Dialogues \\ 
\citet{Kluknavská_2015} & PolSci & Extremist threat & FR:TH & Marian Kotleba, Right-wing Extremism & News \\ 
\citet{goffredo2022fallacious} & AI & Appeal to emotions (fear) & FA:AE & Foreign policy, Economy, Immigration, Healthcare, Social issues & Political Debate \\ 
\citet{prahallad2025analyzing} & CL & Appeal to fear & PR:AF & Immigration, Inflation, National Security, Energy policy, Crime \& law enforcement, Executive Experience \& Leadership & Political Debate \\ 
\citet{jung2024multi} & SI & Security \& Defense & FR:SE & 2019 Quran burning in Norway & Twitter, YouTube, News \\ 
\citet{johnson-etal-2017-leveraging} & CL & Security \& Defense & FR:SE & Abortion, Healthcare, Gun policy, Immigration, Terrorism, LGBTQ & Twitter \\ 
\end{longtable}

\clearpage
\twocolumn

\end{document}